\documentclass{article}

\usepackage{microtype}
\usepackage{graphicx}
\usepackage{caption}
\usepackage{subcaption}
\usepackage{svg}
\usepackage{booktabs} 
\usepackage{tabularx}
\graphicspath{{./images/}{.}}
\makeatletter
\def\input@path{{./images/}{./}}
\makeatother
\usepackage{pgf}
\usepackage{layouts}

\usepackage{svg}
\usepackage{times}
\usepackage{epsfig}
\usepackage{graphicx}
\usepackage{amsmath}
\usepackage{amssymb}
\usepackage{inputenc}
\usepackage{algorithm}
\usepackage{adjustbox}
\usepackage{listings}

\usepackage{amsmath,amsfonts,bm}

\def\eqref#1{equation~\ref{#1}}

\def\1{\bm{1}}

\def\rx{{\textnormal{x}}}

\def\rvp{{\mathbf{p}}}

\def\rvx{{\mathbf{x}}}

\def\rvz{{\mathbf{z}}}

\DeclareMathAlphabet{\mathsfit}{\encodingdefault}{\sfdefault}{m}{sl}
\SetMathAlphabet{\mathsfit}{bold}{\encodingdefault}{\sfdefault}{bx}{n}

\newcommand{\E}{\mathbb{E}}

\newcommand{\R}{\mathbb{R}}

\newcommand{\KL}{D_{\mathrm{KL}}}

\usepackage{xcolor,colortbl}
\definecolor{green}{rgb}{0.91764706, 0.94117647, 0.86666667}
\definecolor{red}{rgb}{0.94509804, 0.85490196, 0.85490196}

\usepackage{hyperref}
\usepackage{mathtools}

\usepackage[preprint]{icml2021}

\icmltitlerunning{Stabilizing VAEs}

\begin{document}

\twocolumn[
\icmltitle{Re-parameterizing VAEs for stability}




\begin{icmlauthorlist}
\icmlauthor{David Dehaene}{to}
\icmlauthor{Rémy Brossard}{to}

\end{icmlauthorlist}

\icmlaffiliation{to}{AnotherBrain, Paris, France}

\icmlcorrespondingauthor{David Dehaene}{david.dehaene@polytechnique.edu}

\icmlkeywords{Machine Learning, Variational Autoencoder}

\vskip 0.3in
]



\printAffiliationsAndNotice{}  

\begin{abstract}
    We propose a theoretical approach towards the training numerical stability of Variational AutoEncoders (VAE).
    Our work is motivated by recent studies empowering VAEs to reach state of the art generative results on complex image datasets.
    These very deep VAE architectures, as well as VAEs using more complex output distributions, highlight a tendency to haphazardly produce high training gradients as well as NaN losses.
    The empirical fixes proposed to train them despite their limitations are neither fully theoretically grounded nor generally sufficient in practice.
    Building on this, we localize the source of the problem at the interface between the model's neural networks and their output probabilistic distributions.
    We explain a common source of instability stemming from an incautious formulation of the encoded Normal distribution's variance, and apply the same approach on other, less obvious sources.
      We show that by implementing small changes to the way we parameterize the Normal distributions on which they rely, VAEs can securely be trained.


\end{abstract}

\section{Introduction}
Variational AutoEncoders (VAE) \citep{kingma2014autoencoding}) are well known generative models used to learn the distribution of the training data. This grants them use in contexts of generative applications \citep{vahdat2020nvae} as well as industrial applications of visual quality control \citep{dehaene2020anomaly}.
Recent works \citep{vahdat2020nvae, verydeepvae}, using hierarchical architectures of latent distributions, showed that VAE could reach state-of-the-art generative and distribution modelling results on complex datasets such as FFHQ \citep{karras2019stylebased} while being generally easier to train than other generative models such as Generative Adversarial Networks (GAN, \cite{goodfellow2014generative}).

Nonetheless, both teams report instability issues during training.
\cite{vahdat2020nvae} mentions the "unbounded KL term" (Kullback Leibler regularization loss reaching infinite values) as a reason for instabilities, and proposes residual Normal distributions and spectral regularization to circumvent them.
\cite{verydeepvae} write \begin{quote}
VAEs have notorious “optimization difficulties,” which are not frequently discussed in the literature but nevertheless well-known by practitioners.  These manifest as extremely high reconstruction or KL losses and corresponding large gradient magnitudes, which may be due to the variance involved in sampling latent variables."
\end{quote}
They observe that cases leading to high gradient values are relatively rare and propose to simply skip their updates as they arise.

Other stability issues arise when VAEs output both mean and variance values parameters for their output distribution, as reported in \cite{lin2019balancing, rybkin2020simple}.
Here, the standard solution is to use a global variance parameter that is independent from the latent space.

In this paper, we will investigate the reasons behind these instabilities, and show very simple changes to the vanilla VAE model that negate their effect.
These changes are theoretically motivated in contrast to previous solutions that were more empirical.
We will show that previous unstable models, augmented with these changes, become stable during training.

\section{Variational AutoEncoders today}
VAEs build on traditional autoencoders, with the goal of forming low dimensional representations of input data. They are composed of a decoder network, with input $\rvz \in \R^L$ and output $\mu_{\theta}(\rvz) \in \R^d, L < d$, with $\theta$ representing the decoder's weigths, and an encoder network, with input $\rvx \in \R^d$ and outputs $\mu_{\phi}(\rvx) \in \R^L$ and $\sigma_{\phi}(\rvx)$ with $\phi$ representing the encoder's weigths.
VAEs \citep{kingma2014autoencoding} add a probabilistic setting to this framework: the data is sampled from a unknown distribution $q(\rvx)$, and the objective is to maximize $\E_{q(\rvx)} p_{\theta}(\rvx)$ with
\begin{equation}
\begin{split}
    p_{\theta}(\rvx) &\coloneqq \E_{p(\rvz)} p_{\theta}(\rvx|\rvz) \\
    p(\rvz) &\coloneqq \mathcal{N}(0, 1)^L \\
    p_{\theta}(\rvx|\rvz) &\coloneqq \mathcal{N}(\mu_{\theta}(\rvz), \gamma)
\end{split}{}
\end{equation}
$\gamma$ was originally defined as a constant, but \cite{TwoStageVAE} proposed to define $\gamma$ as a global learnable parameter to reach a better fit.

The above objective is intractable, so an importance sampling distribution is defined  \begin{equation}
    q_{\phi}(\rvz|\rvx) \coloneqq \mathcal{N}(\mu_{\phi}(\rvx), \sigma_{\phi}(\rvx))
\end{equation}{}
so that a tractable lower bound of our objective can be derived, called the Evidence Lower Bound Objective (ELBO):
\begin{equation}
\begin{split}
        \E_{q(\rvx)} p_{\theta}(\rvx) \geq  \E_{q(\rvx)} [ \, &E_{q_{\phi}(\rvz|\rvx)} \log p_{\theta} (\rvx|\rvz) \\&-  \KL(q_{\phi} (\rvz|\rvx), p(\rvz)) \, ]
\end{split}{}
\end{equation}{}
where KL denotes the Kullback-Leibler divergence.

\citet{Rezende2018GeneralizedEW} showed that the optimization of the $\gamma$ parameter has a closed form solution in the mean squared reconstruction error:
\begin{equation}
    \gamma*^2 = \E_{q(\rvx)} \E_{q(\rvz|\rvx)} \frac{1}{d} \sum_{i=1}^d (\, \mu_{\theta}(\rvz)_i - \rx_i)^2
\label{eq:optimalgamma}
\end{equation}{}

Attempts were made by \citet{lin2019balancing, rybkin2020simple} to replace the global $\gamma$ parameter by a more powerful model $p_{\theta}(\rvx|\rvz) = N(\mu_{\theta}(\rvz), \gamma_{\theta}(\rvz))$, but this endeavour is described as "extremely challenging" \citep{lin2019balancing} and "leading to poor results [because] the network is able to predict certain pixels with very high certainty, leading to degenerate variances." \citep{rybkin2020simple}

Recently, \cite{verydeepvae, vahdat2020nvae} obtained impressive generative results building on hierarchical VAEs \citep{NIPS2016_ddeebdee, sonderby2016ladder}, models that use several levels of latent variables $\rvz_1 ... \rvz_n$ to model the input distribution as a more complex one:
\begin{equation}
    p_{\theta}(\rvx, \rvz_1, ..., \rvz_n) = p(\rvz_n) p_{\theta}(\rvx|\rvz_{\geq 1}) \prod_{i = n..2} p_{\theta}(\rvz_{i-1}|\rvz_{\geq i}) 
\end{equation}{}

\section{Proposed method}
\subsection{Problem analysis}
\cite{vahdat2020nvae} observe that terms in the form of a KL divergence $\KL(q_1, q_2)$ will yield infinite values if the two distributions are very far from each other and call this effect the "Unbounded KL term" problem.
They propose to limit how far the two distributions can drift apart by adding residual Normal distributions and Lipschitz constraints on models parameterizing those distributions, but these empirical solutions don't fully get rid of instabilities.

This KL term being directly minimized as a part of the VAE loss, the gradients should in principle suffice for it to stay bounded during training, but the existence of regions of explosive gradients seem to prevent stable training.
We propose to investigate this problem by analyzing the behavior of the KL for small and large variance values for $q_1$ and $q_2$.
We will first correct a common numerical instability that is bound to occur for a small variance when naively using standard tensor probability frameworks such as TensorFlow Probabilities or PyTorch Distributions.
It will lead us to examine possible constraints to apply to the VAE's distributions, and propose more stable parameterizations, thus eliminating possible explosive gradients.

\subsection{Stabilizing the encoded distribution variance}
\subsubsection{Small encoded distribution variance}
\label{sec:naive_exp}

    Let us explicit the components of the KL term:
\begin{equation}
    \KL(q_1, q_2) = \log \frac{\sigma_2}{\sigma_1} + \frac{\sigma_1^2 + (\mu_1 - \mu_2)^2}{2 \sigma_2^2} - \frac{1}{2}
\label{eq:explicit_kl}
\end{equation}{}
with $q1 = \mathcal{N}(\mu_1, \sigma_1), q2 = \mathcal{N}(\mu_2, \sigma_2)$.
We indeed see that without bounds on $\sigma_1$ and $\sigma_2$ the term $\frac{\sigma_2}{\sigma_1}$ can be so small that $\log \frac{\sigma_2}{\sigma_1}$ can be numerically evaluated to $- \infty$.
An unfortunate accumulations of design choices make this case far less rare than imagined, as summed up in  Table \ref{tab:examples_values}.
First, we noted that this effect is made even worse by some common implementations (ex: the PyTorch implementation) of this KL divergence, where, for speed optimizations, the quotient term $\log \frac{\sigma_2}{\sigma_1}$ is evaluated as $ 0.5*\log\frac{\sigma_1^2}{\sigma_2^2}$, where $\log\frac{\sigma_1^2}{\sigma_2^2}$ will collapse numerically to $-\infty$ for even higher, more typical, values of $\frac{\sigma_2}{\sigma_1}$.
Second, a common practice to generating the positive scale parameter for the Normal distribution is to exponentiate some parameter neural network output $p_i$: $\sigma_i = \exp(p_i)$. 
This makes it even easier to fall in the divergent zone, with a small shift in $p_i$.
Finally, for memory and speed optimizations, recent models use half precision floating point arithmetic, which will worsen this effect.

\begin{table}[h]
\caption{Minimal values of $\sigma_1$ and $\widehat{\sigma_1} = \log \sigma_1$ for $\log \frac{\sigma_2}{\sigma_1}$ and $0.5*\log\frac{\sigma_1^2}{\sigma_2^2}$ (PyTorch) implementations, in the NaiveExp parameterization and for $\sigma_2 = 1$, for different floating point precisions.}
\label{tab:examples_values}
\vskip 0.15in
\begin{tabularx}{\linewidth}{@{}lXXXX@{}}
\toprule
        & min($\widehat{\sigma_1}$) & min($\sigma_1$)       &
        \begin{tabular}{c}min($\widehat{\sigma_1}$) \\ PyTorch\end{tabular} &  \begin{tabular}{c}min($\sigma_1$) \\ PyTorch\end{tabular}   \\ \midrule
float32 & -103   & 1.85e-45 & -51    & 7.10e-23 \\
float16 & -17    & 4.14e-08 & -8     & 3.35e-04 \\ \bottomrule
\end{tabularx}
\end{table}

For future reference, we will call NaiveExp the above parameterization
\begin{equation}
\begin{split}{}
    \sigma_{ NaiveExp }(p) &= \exp(p) \\
    \widehat{\sigma}_{ NaiveExp }(p) &= \log(\exp(p))
    \end{split}
\end{equation}{}
with each operation being numerically evaluated, so that $\log(\exp(p)) \ne p$.

The solution to this effect is quite simple.
First, let us define $\widehat{\sigma_1} = \log \sigma_1, \widehat{\sigma_2} = \log \sigma_2$, so that \eqref{eq:explicit_kl} becomes
\begin{equation}
    \KL(q_1, q_2) = \widehat{\sigma_2} - \widehat{\sigma_1} + \frac{\sigma_1^2 + (\mu_1 - \mu_2)^2}{2 \sigma_2^2} - \frac{1}{2}
\end{equation}{}

Remember that computational problems come from the explicit and not simplified computations $\log(\exp(p))$ 
(by using popular libraries PyTorch Distributions and TensorFlow Probability, it is easy to inadvertedly use NaiveExp).
By instead using the following parameterization, that we will call Exp:
\begin{equation}
\begin{split}{}
    \sigma_{ Exp } (p)  &= \exp(p) \\
    \widehat{\sigma}_{ Exp } (p) &= p
    \end{split}
\label{eq:Exp}
\end{equation}{}
the KL becomes
\begin{equation}
\begin{split}
      \KL(q_1, q_2) = &\frac{1}{2} (\exp(2 p_1) + (\mu_1 - \mu_2)^2) \exp(- 2 p_2) \\ &+ p_2 - p_1  - \frac{1}{2} 
\end{split}
\label{eq:exp_param}
\end{equation}{}
(Remember that exponential terms were simply made explicit here, and were always present behind the generation of $\sigma$ as a positive term).
We can see that in this expression, we no longer have to fear an infinite term for $\sigma_1 \ll \sigma_2$, which would have been retropropagated as NaN in the gradients computations.

\subsection{Large encoded distribution variance}
Using Exp, in \eqref{eq:exp_param}, large gradients can still be produced by high values of $p_1$, going through  exponentiation.
To prevent such problems, we propose two other parameterizations for Normal distributions.
The first one, ExpLin, reduces the increase rate for large values of $\sigma$ by replacing the exponential by a linear function:
\begin{equation}
\begin{split}{}
    \sigma_{ ExpLin } (p)  &= 
    \begin{cases} \exp(p) &\text{ if } p \leq 0 \\   p + 1 &\text{ if } p > 0
    \end{cases} \\
    \widehat{\sigma}_{ ExpLin } (p)  &= 
    \begin{cases} p &\text{ if } p \leq 0 \\ \log(p+1) &\text{ if } p > 0 
    \end{cases}
    \end{split}
\end{equation}{}

We also more radically propose UpBounded$\omega$, where we cap the value of the scale to a constant $\omega$
\begin{equation}
\begin{split}{}
    \sigma_{ UpBounded\omega }(p) &=
    \begin{cases} \frac{\omega}{2}\exp(p) &\text{ if } p \leq 0  \\
    \frac{\omega}{2} (2 - \exp(-p)) &\text{ if } p > 0
    \end{cases} \\
    \widehat{\sigma}_{ UpBounded\omega } (p) &= 
    \begin{cases} p + \log \frac{\omega}{2} &\text{ if } p \leq 0 \\ \log(2 - \exp(-p)) + \log \frac{\omega}{2} &\text{ if } p > 0
    \end{cases}
    \end{split}
    \label{eq:upbounded}
\end{equation}{}
Typically setting $\omega = 1$.
This particular value is justified because, without the reconstruction term in the loss, the optimal encoded distribution scale would be 1, fitting the prior distribution $p(\rvz) = \mathcal{N}(0, 1)^L$. 
If the VAE encodes information using its encoded mean, then the higher $\sigma_1$ is, the higher the reconstruction error will be, so that the presence of the reconstruction term decreases $\sigma_1$, and $0 \leq \sigma_1 \leq 1$.

\subsection{Parameterized priors}
After examining the case of encoded distributions, affecting the left hand side of the KL term, let's focus on the right hand side parameter $\sigma_2$. 
In the original formulation of VAEs, $\sigma_2$ has a fixed value of 1, but in the case of autoregressive models such as NVAE and VDVAE, the prior part of the KL computation can also be output by the model.
In that case, we cannot set an upper bound to $\sigma_1$ to ensure that it is smaller than $\sigma_2$, as above.
We propose to also use the UpBounded$1$ parameterization for \emph{decoded} distributions in latent spaces. 
We will also present results using the ExpLin parameterization.
Finally, we put forward a slightly heavier modification of the standard autoregressive VAE framework: setting $\sigma_2 = 1$ as a constant value for all latent spaces.
Indeed, theoretically, since in latent spaces values are both encoded and decoded by models, a constant variance value from the decoder can be compensated on the encoder's side. 
If the parameterization of the variance is the unstable element in the training of a VAE, this constant value should help stabilize our models.

\section{Stabilizing the decoded distribution variance}
\label{sec:decoded}
The same parameterizations of the Normal distribution can be used for the decoded distribution $p_{\theta}(\rvx|\rvz)$.
Namely, UpBounded$\omega$ can be fitted to the range of the input data.
For data in [0, 256[, such as 8 bit images, an upper bound to the decoder standard deviation is 256 (reached for a perfectly wrong decoder, using \eqref{eq:optimalgamma}), so we can set $\omega = 256$.

An issue that is often noted concerning the term $\gamma_{\theta}(\rvz)$ is that is can converge very fast towards infinitely small values for pixels that are easy to reconstruct \citep{lin2019balancing, rybkin2020simple}.
This in turns leads to even more incentive for these pixels to have a better reconstruction, applied by the reconstruction term 
\begin{equation}
    \frac{(\mu_{\theta}(\rvz)_i - \rx_i)^2}{\gamma_{\theta}(\rvz)_i^2}
\end{equation}{}
for some $\rvx, \rvz$, and $i$ the index of one of these pixels, which has more weight compared to other pixels, the more $\gamma_{\theta}(\rvz)^2_i$ is small.
This vicious cycle leads to instability in the training of the decoded distribution.

But the fact that this vicious cycle leads to infinitely small values of $\gamma_{\theta}(\rvz)$ is a byproduct of the continuous nature of our model, applied on discrete data.
\citet{theis2016note} note that to properly evaluate continuous models on discrete data, one has to add uniform noise to the data.
For example, if the inputs are in \{0, 1.. 255\}, uniform noise must be sampled from [0, 1[ and added to the inputs for a fair evaluation of the model.
This naturally leads to a lower bound for $\gamma_{\theta}(\rvz)$, called $\alpha$.
We use \eqref{eq:optimalgamma} to set $\alpha$ to $\left[ \, \int_{[0,1[} (x - \frac{1}{2})^2 dx \, \right]^{0.5} = \frac{1}{\sqrt{12}}$ for uniform [0, 1[ noise as above.

This leads to the DownBounded$\alpha$ parameterization:
\begin{equation}
\begin{split}{}
    \sigma_{ DownBounded\alpha }(p) &= 
    \alpha + \exp(p) \\
    \widehat{\sigma}_{ DownBounded\alpha } (p) &= \log \sigma_{ DownBounded\alpha }(p)
    \end{split}
\end{equation}{}
And, combined with UpBounded, it yields the Bounded parameterization, which is the most natural for our decoded distribution:
\begin{equation}
\begin{split}{}
    \sigma_{ Bounded\alpha\omega }(p) &= 
    \alpha + (\omega - \alpha)  s(p) \\
    \widehat{\sigma}_{ Bounded\alpha\omega } (p) &= \log \sigma_{ Bounded\alpha\omega }(p)
    \end{split}
\end{equation}{}
with $s$ the sigmoid function.
Notice that the use of the sigmoid is an arbitrary choice among smooth bounded functions.

\section{Note on software implementation}

As the implementations of the conditional parameterizations themselves are
prone to creating NaNs, we suggest here a possible working and fast implementation of, for example, ExpLin.
Random access indexing on large tensors being slow, we use a multiplicative mask:
\begin{equation}
    \sigma_{ExpLin} = e^\rvp * \delta(\rvp \leq 0) + (\rvp + 1) * \delta(\rvp > 0)
\label{eq:forward}
\end{equation}{}
where $\delta(condition) = 1 \text{ if } condition \text{ is true else} = 0$.
However, $e^{p_i}$ is still evaluated for values $p_i > 0$, even though the result is multiplied by 0.
For high values of $p_i$, so that $e^{p_i}$ is numerically evaluated to $+\infty$, we get     \begin{equation}
    \sigma_{ExpLin_i} = +\infty * 0 + (p_i + 1) * 1 = NaN
\end{equation}{}

We found that the $Clip$ function is well optimized, so our solution is to compute two clipped versions of the $\rvp$ tensor, one clipped to 0 for $p_i > 0$ and the other clipped to 0 for $p_i < 0$. Conditional parameterizations are then applied to the clipped versions of $\rvp$.
Thus, we replace \eqref{eq:forward} with
\begin{equation}
    \begin{split}
    \rvp^+ &= ClipMin(\rvp, 0) \\
    \rvp^- &= ClipMax(\rvp, 0) \\
    \sigma_{ExpLin} = e^{\rvp^-} * \delta(&\rvp^- \leq 0) + (\rvp^+ + 1) * \delta(\rvp^+ > 0)
    \end{split}{}
\end{equation}{}

\section{Related Work}

The original VAE formulation \citep{kingma2014autoencoding} only required positivity of the different variances of the model, which was ensured setting them as the result of an exponential. This is the first specific constraint found in litterature.

Very few work report training instabilities for VAEs in the next years, as they are viewed as the more stable but less expressive counterpart to GANs \citep{goodfellow2014generative} or BiGANs \citep{donahue2017adversarial}, which are roughly GANs with encoder networks. 
Recently however, NVAE \citep{vahdat2020nvae} and VDVAE  \citep{verydeepvae} changed this point of view. They are powerful generative VAE models, unfortunately coming with their share of training instabilities. 

Two different strategies are used to reduce these instabilities: first, \citet{verydeepvae} suggest modifications to the model's weight initialization and show, given the right hyperparameters, that the instabilities are sufficiently sparse to be simply ignored during training, by skipping updates of NaN or high gradient. Then, \citet{vahdat2020nvae} propose empirical constraints to their models to reduce the probability of unstable gradients.
They propose Residual Normal distributions, distributions that are parameterized using a differential mean $\Delta\mu$ from the prior's mean, and a differential scale $\Delta\sigma$ from the prior's scale.
Since in the context of their model the prior and the encoded distributions are both trained concurrently, this parameterization is supposed to be easier than an absolute parameterization.
They also propose Spectral Regularization, an additional term in the loss to promote smoothness in the distributions parameters.
While these improvements enabled them to train impressive models, NaNs can still randomly happen during training.

A source of instabilities that has been identified in previous work is the decoded variance, that is to say $p_{\theta}(\rvx|\rvz) = N(\mu_{\theta}(\rvz), \gamma_{\theta}(\rvz))$, making the variance of the decoded distribution an output of the decoder model.
However, this parameterization is seen as  "extremely challenging" \citep{lin2019balancing} and a more cautious parameterization using a learnable but global variance parameter $p_{\theta}(\rvx|\rvz) = N(\mu_{\theta}(\rvz), \gamma_G)$, as first proposed in \citet{TwoStageVAE}, is almost always preferred.
\citet{lin2019balancing} propose a two-staged approach to stably train the decoded variance distribution: first they learn a global parameter $\gamma_G$ as above.
Then, they constrain the more powerful decoded variance $\gamma_{\theta}(\rvz)$ to be greater than $\frac{\gamma_G} {\alpha}$ with $\alpha$ being an hyperparameter.

\section{Experiments}

\begin{table}[htb]
\caption{Proportion of models trained on only 10 images that returned NaN result or failed to converge, for different learning rates (LR).}%
\label{results: overfitted VAE - latent and decoder parametrization}
\vskip 0.15in
\centering%
\begin{tabularx}{\linewidth}{@{}lXXXXXXX@{}}
\toprule%
LR &Encoded~$\sigma$ Param.&\multicolumn{3}{c}{Decoder $\gamma$ Parametrization}\\%
\midrule
&&NaiveExp&Exp&Bounded\\%
\midrule%
0.008&NaiveExp&10 / 10&10 / 10&10 / 10\\%
~&Exp&10 / 10&7 / 10&8 / 10\\%
~&ExpLin&0 / 10&0 / 10&0 / 10\\%
~&UpBounded&0 / 10&0 / 10&0 / 10\\%
\midrule%
0.007&NaiveExp&10 / 10&8 / 10&9 / 10\\%
~&Exp&4 / 10&7 / 10&3 / 10\\%
~&ExpLin&0 / 10&0 / 10&0 / 10\\%
~&UpBounded&0 / 10&0 / 10&0 / 10\\%
\midrule%
0.006&NaiveExp&7 / 10&9 / 10&3 / 10\\%
~&Exp&5 / 10&6 / 10&2 / 10\\%
~&Explin&0 / 10&0 / 10&0 / 10\\%
~&UpBounded&0 / 10&0 / 10&0 / 10\\%
\bottomrule%
\end{tabularx}
\end{table}


\subsection{Stabilizing the latent space of overfitted model}

NVAE and VDVAE being very expensive models to train, we set out to find a simplified experiment exhibiting training instabilities (experiments on the full VDVAE will be developed in section \ref{sec:vdvae}). 
Notably, instabilities seem to happen if some part of the
dataset is overfitted, especially if the learning rate is high.
We emulate this behavior to assert that the proposed correction fixes this type of instability issue. We use a subset of CIFAR-10 with only 10 different images as the training dataset. The decoder variance is parameterized by a single learnable scalar.

We compare VAEs with different parameterizations of the encoded and decoder variance. 
We trained 10 versions of each model with different learning rates for 20000 batches of 512 samples. Each training takes approximately 12 minutes on a 1080 Ti GPU. In Table \ref{results: overfitted VAE - latent and decoder parametrization}, we show the number of each type of model that produced NaN values or failed to converge (having negative ELBOs orders of magnitude higher than normal). No significant difference in attained loss was measured between the different converged models.

We observe that models with a naive parameterization of the encoded variance are extremely unstable in that setting, and that most training experiments will fail.
However, all models with the encoded variance parameterized as either ExpLin or UpBounded converged.

One the other hand, the parameterization of the decoder variance seems less critical in this setup.



\subsection{Parameterization of the decoded variance}

In this subsection we will study a simple VAE where the encoder and decoder are each composed of two fully connected linear layers.
The hidden layer size is 128, and the number of latent dimensions is 128.
We will apply it on the MNIST dataset \citep{mnist}. We separated the given training set between a training set (first 50000 samples) and validation set (10000 remaining samples). MNIST has the interesting property that certain pixels have a constant value over the whole dataset (e.g. corners are 0-valued in every image).
This facilitates the work of the decoder, that should be able to output a precise distribution $p(\rx_{corner}|\rvz) = \mathcal{N}(0, \gamma_{corner})$ with $\gamma_{corner}$ arbitrarily small.
As this experiment is made to show limitations of the VAE in such a case, noise will only be added to the data, as in \citet{theis2016note}, for the evaluation of the NLL. 
According to section \ref{sec:decoded}, VAE with a decoded variance should be unstable on this dataset without our added parameterizations, because $\gamma_{corner}$ will reach the limits of the floating point precision.

We set the learning rate to 0.0001 and train the models with 5 different initializations, for 1000 epochs. The best model in validation Negative Log Likelihood is tested.
The whole experiment takes approximately 10 hours on a 1080Ti GPU.
The mean of the decoded distribution is either directly output by the decoder, or passed through a sigmoid activation to restrict it to values in [0, 1], the range of MNIST values.
The encoded distribution is parameterized using UpBounded1.
The decoded distribution is parameterized using either NaiveExp, Exp, UpBounded1,  DownBounded$\alpha$, or Bounded$\alpha$1 with $\alpha = \frac{1}{256 \sqrt{12}}$. 
The last tested option for the decoded distribution is to be parameterized by a global scalar parameter, common for all pixel positions and all inputs, combined with NaiveExp. 
We call this last one Global.

\begin{table}[h]

\centering
\caption{MNIST Negative Log Likelihoods for different decoded distributions parameterizations.}
\vskip 0.15in
\begin{tabularx}{\linewidth}{@{}lXX@{}}
\toprule
        & $\mu_{identity}$ & $\mu_{sigmoid}$  \\    \midrule

$\gamma_{NaiveExp}$ & 5.35 $\pm$ 0.05   & 100\% NaN  \\
$\gamma_{Exp}$ & 5.36 $\pm$ 0.06 & 100\% NaN \\
$\gamma_{UpBounded}$ & 5.25 $\pm$ 0.02 & 100\% NaN \\
$\gamma_{DownBounded}$ & 5.30 $\pm$ 0.06 & \textbf{4.56} $\pm$ 0.07 \\
$\gamma_{Bounded}$ & 5.05 $\pm$ 0.07    & \textbf{4.56} $\pm$ 0.05  \\
$\gamma_{Global}$ & 6.67 $\pm$ 0.004     & 6.06 $\pm$ 0.02
\\ \bottomrule
\end{tabularx}
\label{tab:MNIST_exp}

\end{table}

We can see in Table \ref{tab:MNIST_exp} that even though Global is stable, as noted in previous work, decoding the variance leads to better fitting models.
In the case of $\mu_{sigmoid}$, both NaiveExp, Exp and UpBounded yield NaN results independently of the initialization.
Contrary to the encoded distribution, fixing the NaiveExp parameterization for small variances is not enough in this case because the decoded variance reaches a perfect 0, simulating a Dirac distribution, so that e.g. for Exp:
\begin{equation}
\begin{split}{}
    \sigma_{ Exp } (-\infty)  &= \exp(p) = 0 \\
    \widehat{\sigma}_{ Exp } (p) &= p = -\infty
    \end{split}
\end{equation}{}
However, both parameterizations having a minimum variance, DownBounded and Bounded, are stable and give the best results.

\begin{figure}[!h]
\input{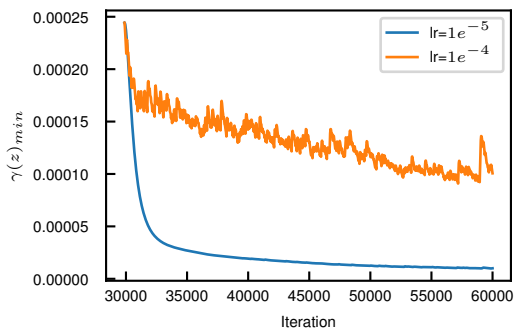}
\caption{$\gamma(\rvz)_{min}$ for a $\gamma_{NaiveExp}, \mu_{identity}$ model pre-trained for 1000 epochs with lr=1e-4, then trained for 1000 epochs with either lr=1e-4 or lr=1e-5. We see that  the maximal precision of the model, reflected by $\gamma(\rvz)_{min}$, depends heavily on the learning rate. For $\mu_{sigmoid}$, this precision becomes infinite for the 0 target, so that a $\gamma_{NaiveExp}$ model diverges.}
\label{fig:lr}
\end{figure}

Interestingly, the model does not diverge using $\mu_{identity}$. 
We believe that is because the decoder has troubles decoding a perfect 0 in this setting, as hinted by the minimum of the decoded variances, that seem to reach a limit depending on the learning rate (Figure \ref{fig:lr}).
In the $\mu_{sigmoid}$ case, however, this difficulty is removed as the decoder has a much easier time targetting 0 by setting the input of the sigmoid to $- \infty$, and such models yield NaN as predicted.

As we are seeking a general stable framework, independently of the model's capacity or its learning rate, this experiment strongly promotes the use of Bounded or DownBounded as the decoded distribution's parameterization.

\begin{figure}[!h]
\input{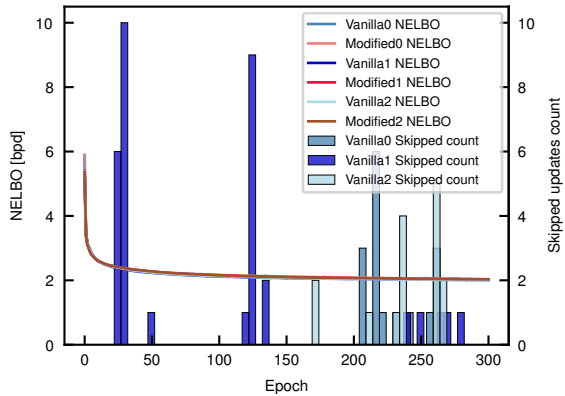}
\caption{Negative ELBO in bits per dimension and number of skipped updates in VDVAE training, comparing the version from \citet{verydeepvae} (Vanilla) and our Modified BoundedExp1 version on several seeds. Vanilla models are shown in blue, and modified models are shown in red. The skipped updates count for modified models is 0.}
\label{fig:nan}
\end{figure}

\subsection{Stabilization of Very Deep VAEs}
\label{sec:vdvae}

\citet{verydeepvae} provided public access to the code for their Very Deep VAE.
Running the experiment with the provided hyperparameters on the CIFAR10 \citep{cifar10} dataset, we run into several NaN values both in the KL and reconstruction parts of the ELBO. 
When a NaN value or high gradients appear during training, their code skips the current update.
We compared the base implementation of the model (Vanilla) with an implementation where we replaced every parameterization of a Normal distribution (encoded or decoded) with the UpBounded1 parameterization (Modified).
The decoded distribution in pixel space remains the Multivariate Discretized Logistic distribution of the original implementation.
Both versions of the model were run three times, with seed 0 to 2, on two 16-GB v100 GPUs.
Each training lasts approximately 3 days, for 300 epochs.

\begin{figure}[h]
\centering
\includegraphics[scale=0.8]{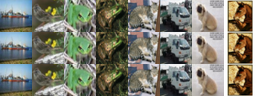}
\includegraphics[scale=0.43]{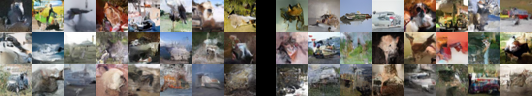}
\caption{Top: Samples from CIFAR10 (top), reconstructed from a Vanilla VDVAE (middle) and a modified VDVAE (bottom).
Bottom: Generated samples from a Vanilla VDVAE (left) and a modified VDVAE (right)}
\label{fig:reconstructions}
\end{figure}

The reconstructions and generations from Figure \ref{fig:reconstructions} as well as the test negative ELBO in Table \ref{tab:nan_count} show that the performance of the model is not influenced by our modifications. 
Figure \ref{fig:nan} shows training curves and skipped update events of each models.
The total number of NaN events are also reported in Table \ref{tab:nan_count}.
It can be seen that no high gradient or NaN event were encountered during training of the modified models.

\begin{table}[h]
\small
\centering
\caption{Number of high gradient skipped updates and number of NaN encountered during training and test Negative ELBO in bits per dimension (lower is better)}
\vskip 0.15in
\begin{tabularx}{\linewidth}{@{}lXll@{}}
\toprule
& High gradients count & NaN count & Test NELBO\\    \midrule
Vanilla & 22 $\pm$ 9 & 13 $\pm$ 3   & 2.014 $\pm$ 0.009  \\
UpBounded1 & \textbf{0 $\pm$ 0} & \textbf{0 $\pm$ 0} & 2.020 $\pm$ 0.006  \\
ExpLin & 47 $\pm$ 61 & \textbf{0 $\pm$ 0} & 2.030 $\pm$ 0.011  \\
Vanilla/Cst1 & 37 $\pm$ 43 & 14 $\pm$ 16 & 2.009 $\pm$ 0.003  \\
UpBounded1/Cst1 & 2645 $\pm$ 2063 & \textbf{0 $\pm$ 0} & 2.028 $\pm$ 0.007 \\
ExpLin/Cst1 & 1 $\pm$ 1 & \textbf{0 $\pm$ 0} & \textbf{2.006 $\pm$ 0.008} \\
\bottomrule
\end{tabularx}
\label{tab:nan_count}
\end{table}

\subsection{Parameterization study of VDVAE}
For completeness, we also present in Table \ref{tab:nan_count} the results of parameterizing every Normal distribution in VDVAE with ExpLin.
We see that even though a few high gradient values are encountered during training, no NaN occur with this parameterization.

Finally, we examine the case where the variance of the Normal distributions in the decoder path are all set to a constant 1.
Encoded distributions are either the Vanilla version from the original VDVAE implementation (Vanilla/Cst1), or our UpBounded1 and ExpLin parameterizations (UpBounded1/Cst1 and ExpLin/Cst1).
We can see that this interestingly does not hinder the performance of the model.
Indeed, this constant value does not theoretically remove expressiveness from the model, as the encoder should be able to appropriately dilate or contract regions of its encoded space to compensate this constraint.
The UpBounded1 case yields a high number of high gradient values during training, but the ExpLin version is both stable and competitive in terms of test NELBO.

This experiment shows that our modifications enable stable convergence of VDVAE models, without requiring the update skipping trick from the original implementation, thus ensuring that the whole dataset is used during training.

\section{Conclusion}

We showed that the original formulation of VAE Normal distributions came with a few traps that could create instability during training.
However, by constraining the interface between the encoder and decoder networks on one side, and the encoded and decoded distributions on the other, we were able to tame these instabilities.
For the encoded distribution, we recommend the UpBounded parameterization corresponding to the variance of the prior, and for the decoded distribution we recommend the Bounded parameterization corresponding to the range of the input data.
For autoregressive models, our work indicates that the value of the decoded variance can be fixed to a constant, and the ExpLin parameterization can be used on the encoded side.
We are confident that using these new parameterizations researchers will be able to fearlessly try more and more powerful architectures for VAEs.

\bibliographystyle{icml2021}
\bibliography{references}

\section*{Acknowledgement}
This work was granted access to the HPC resources of IDRIS (Jean Zay) under GENCI allocation AD011012173.

\onecolumn

\section*{Experiments details}
The following model is used for the experiment in 7.1:
\begin{lstlisting}{}
VAE(
  (network): Sequential(
    (0): Encoder(
      (main): ConvNetwork(
        (main): Sequential(
          (0): Conv2d(3, 16, kernel_size=(4, 4), stride=(2, 2), padding=(1, 1))
          (1): BatchNorm2d(16, eps=1e-05, momentum=0.1, affine=True)
          (2): ReLU()
          (3): Conv2d(16, 32, kernel_size=(4, 4), stride=(2, 2), padding=(1, 1))
          (4): BatchNorm2d(32, eps=1e-05, momentum=0.1, affine=True)
          (5): ReLU()
          (6): Conv2d(32, 64, kernel_size=(4, 4), stride=(2, 2), padding=(1, 1))
        )
      )
      (mu): Conv2d(64, 100, kernel_size=(4, 4), stride=(1, 1))
      (param_sigma): Conv2d(64, 100, kernel_size=(4, 4), stride=(1, 1))
    )
    (1): Sampler()
    (2): Decoder(
      (to_same_dim): ConvTranspose2d(100, 64, kernel_size=(4, 4), stride=(1, 1))
      (main): ConvNetwork(
        (main): Sequential(
          (0): ConvTranspose2d(64, 32, kernel_size=(4, 4), stride=(2, 2), padding=(1, 1))
          (1): BatchNorm2d(32, eps=1e-05, momentum=0.1, affine=True)
          (2): ReLU()
          (3): ConvTranspose2d(32, 16, kernel_size=(4, 4), stride=(2, 2), padding=(1, 1))
          (4): BatchNorm2d(16, eps=1e-05, momentum=0.1, affine=True)
          (5): ReLU()
          (6): ConvTranspose2d(16, 3, kernel_size=(4, 4), stride=(2, 2), padding=(1, 1))
        )
      )
    )
  )
)
\end{lstlisting}

The training set is composed of 10 random images from the CIFAR10 dataset \citep{cifar10}. 
The images are randomly selected at the beginning of each training.
Input and output images are of dimensions 3x32x32.
The number of latent dimensions is 100.

The following model is used for the experiment in 7.2:

\begin{lstlisting}
BasicVAE(
  (encoder): Sequential(
    (0): Linear(in_features=784, out_features=128, bias=True)
    (1): ReLU()
    (2): Linear(in_features=128, out_features=256, bias=True)
  )
  (decoder): Sequential(
    (0): Linear(in_features=128, out_features=128, bias=True)
    (1): ReLU()
    (2): Linear(in_features=128, out_features=1568, bias=True)
  )
)
\end{lstlisting}{}
The dataset used is MNIST \citep{mnist}. Input and output images are of dimensions 1x28x28. The number of latent dimensions is 128.

For both experiments in 7.1 and 7.2, we used the Adam optimizer \citep{kingma2017adam}.

For the experiment in 7.3, we used the code from \url{https://github.com/openai/vdvae}, executed with the following command:
\begin{lstlisting}[language=bash]
python train.py --hps cifar10 --num_epochs 300 
\end{lstlisting}{}

\end{document}